\documentclass{article}

\usepackage[T1]{fontenc}    
\usepackage[space]{grffile}
\usepackage[utf8]{inputenc} 
\usepackage{algorithm}
\usepackage{algorithm}
\usepackage{algpseudocode}
\usepackage{algpseudocode}
\usepackage{amsfonts}       
\usepackage{amsmath}
\usepackage{amssymb}
\usepackage{amsthm}
\usepackage{arxiv}
\usepackage{booktabs}       
\usepackage{color}
\usepackage{graphicx}
\usepackage{hyperref}       
\usepackage{latexsym}
\usepackage{lipsum}		
\usepackage{microtype}      
\usepackage{multirow}
\usepackage{nicefrac}       
\usepackage{times}
\usepackage{url}            
\usepackage{verbatim}

\newtheorem{theorem}{Theorem}

\makeatletter

\providecommand{\tabularnewline}{\\}
\newcommand{\lyxdot}{.}

\makeatother

\begin{document}
  \title{
    A Simple Discriminative Training Method for Machine Translation with
    Large-Scale Features
  }

  \author{
    Tian Xia \\
    Wright State University \\
    \texttt{SummerRainET2008@gmail.com} \\
    \And
    Shaodan Zhai \\
    Wright State University \\
    \texttt{ShaodanZhai@gmail.com} \\
    \And
    Shaojun Wang \\
    Wright State University \\
    \texttt{SWang.USA@gmail.com} \\
  }
  \maketitle

  \begin{abstract}
    Margin infused relaxed algorithms (MIRAs) dominate model tuning in
    statistical machine translation in the case of large scale features,
    but also they are famous for the complexity in implementation. We
    introduce a new method, which regards an N-best list as a permutation
    and minimizes  the Plackett-Luce loss of ground-truth permutations.
    Experiments with large-scale features demonstrate that, the new method
    is more robust than MERT; though it is only matchable with MIRAs,
    it has a comparatively advantage, easier to implement.
  \end{abstract}

  \section{Introduction}

  Since Och \cite{och2003minimum} proposed minimum error rate training
  (MERT) to exactly optimize objective evaluation measures, MERT has
  become a standard model tuning technique in statistical machine translation
  (SMT). Though MERT performs better by improving its searching algorithm
  \cite{macherey2008lattice,cer2008regularization,galley2011optimal,moore2008random},
  it does not work reasonably when there are lots of features\footnote{The regularized MERT seems promising from Galley et al. \cite{galley2013regularized}
  at the cost of model complexity.}. As a result, margin infused relaxed algorithms (MIRA) dominate in
  this case \cite{mcdonald2005online,watanabe2007online,chiang2008online,tan2013corpus,cherry2012batch}.

  In SMT, MIRAs consider margin losses related to sentence-level BLEUs.
  However, since the BLEU is not decomposable into each sentence, these
  MIRA algorithms use some heuristics to compute the exact losses, e.g.,
  pseudo-document \cite{chiang2008online}, and document-level loss
  \cite{tan2013corpus}. 

  Recently, another successful work in large-scale feature tuning include
  force decoding based\cite{yu2013max}, classification based \cite{hopkins2011tuning}.

  We aim to provide a simpler tuning method for large-scale features
  than MIRAs. Out motivation derives from an observation on MERT. As
  MERT considers the quality of only top$1$ hypothesis set, there might
  have more-than-one set of parameters, which have similar top1 performances
  in tuning, but have very different topN hypotheses. Empirically, we
  expect an ideal model to benefit the total N-best list. That is, better
  hypotheses should be assigned with higher ranks, and this might decrease
  the error risk of top1 result on unseen data.

  Plackett\cite{plackett1975analysis} offered an easy-to-understand
  theory of modeling a permutation. An N-best list is assumedly generated
  by sampling without replacement. The $i$th hypothesis to sample relies
  on those ranked after it, instead of on the whole list. This model
  also supports a partial permutation which accounts for top $k$ positions
  in a list, regardless of the remaining. When taking $k$ as 1, this
  model reduces to a standard conditional probabilistic training, whose
  dual problem is actual the maximum entropy based \cite{och2002discriminative}.
  Although Och \cite{och2003minimum} substituted direct error optimization
  for a maximum entropy based training, probabilistic models correlate
  with BLEU well when features are rich enough. The similar claim also
  appears in \cite{zhu2001kernel}. This also make the new method be
  applicable in large-scale features.

  \section{Plackett-Luce Model}

  Plackett-Luce was firstly proposed to predict ranks of horses in gambling
  \cite{plackett1975analysis}. Let $\mathbf{r}=(r_{1},r_{2}\ldots r_{N})$
  be $N$ horses with a probability distribution $\mathcal{P}$ on their
  abilities to win a game, and a rank $\mathbf{\pi}=(\pi(1),\pi(2)\ldots\pi(|\mathbf{\pi}|))$
  of horses can be understood as a generative procedure, where $\pi(j)$
  denotes the index of the horse in the $j$th position.

  In the 1st position, there are $N$ horses as candidates, each of
  which $r_{j}$ has a probability $p(r_{j})$ to be selected. Regarding
  the rank $\pi$, the probability of generating the champion is $p(r_{\pi(1)})$.
  Then the horse $r_{\pi(1)}$ is removed from the candidate pool.

  In the 2nd position, there are only $N-1$ horses, and their probabilities
  to be selected become $p(r_{j})/Z_{2}$, where $Z_{2}=1-p(r_{\pi(1)})$
  is the normalization. Then the runner-up in the rank $\pi$, the $\pi(2)$th
  horse, is chosen at the probability $p(r_{\pi(2)})/Z_{2}$. We use
  a consistent terminology $Z_{1}$ in selecting the champion, though
  $Z_{1}$ equals $1$ trivially.

  This procedure iterates to the last rank in $\pi$. The key idea for
  the Plackett-Luce model is the choice in the $i$th position in a
  rank $\mathbf{\pi}$ only depends on the candidates not chosen at
  previous stages. The probability of generating a rank $\pi$ is given
  as follows

  \begin{equation}
    p(\mathbf{\pi})=\prod_{j=1}^{|\boldsymbol{\pi}|}\frac{p(r_{\pi(j)})}{Z_{j}}\label{eq: full-formula}
  \end{equation}
  where $Z_{j}=1-\sum_{t=1}^{j-1}p(r_{\pi(t)})$.

  We offer a toy example (Table \ref{Tbe:model}) to demonstrate this
  procedure. 

  \begin{table}[h]
    \begin{centering}
      \begin{tabular}{|c|c|c|c|}
        \hline 
        $\mathbf{r}$ & $r_{1}$ & $r_{2}$ & $r_{3}$\tabularnewline
        \hline 
        $\mathbf{\pi}$ & 2 & 3 & 1\tabularnewline
        \hline 
        $Z$ & 1 & 1-$p(r_{2})$ & 1-($p(r_{2})+p(r_{3}))$\tabularnewline
        \hline 
        $p(\pi)$ & $\frac{p(r_{2})}{Z_{1}}$ & $\frac{p(r_{3})}{Z_{2}}$ & $\frac{p(r_{1})}{Z_{3}}$\tabularnewline
        \hline 
      \end{tabular}
      \par
    \end{centering}
    \caption{The probability of the rank $\boldsymbol{\pi}=(2,3,1)$ is $p(r_{2})\cdot p(r_{3})/(1-p(r_{2}))$
    in a simplified form, as $\frac{p(r_{2})}{Z_{1}}=p(r_{2})$ and $\frac{p(r_{1})}{Z_{3}}=1$.}
    \label{Tbe:model}
  \end{table}

  \begin{theorem} 
    The permutation probabilities $p(\mathbf{\pi})$
    form a probability distribution over a set of permutations $\Omega_{\pi}$.
    For example, for each $\mathbf{\pi}\in\Omega_{\pi}$, we have $p(\mathbf{\pi})>0$,
    and $\sum_{\pi\in\Omega_{\pi}}p(\mathbf{\pi})=1$. 
  \end{theorem}

  We have to note that, $\Omega_{\pi}$ is not necessarily required
  to be completely ranked permutations in theory and in practice, since
  gamblers might be interested in only the champion and runner-up, and
  thus $|\mathbf{\pi}|\le N$. In experiments, we would examine the
  effects on different length of permutations, systems being termed
  $PL(|\pi|)$.

  \begin{theorem}
    Given any two permutations $\mathbf{\pi}$ and $\mathbf{\pi}\prime$,
    and they are different only in two positions $p$ and $q$, $p<q$,
    with $\pi(p)=\mathbf{\pi}\prime(q)$ and $\pi(q)=\mathbf{\pi}\prime(p)$.
    If $p(\pi(p))>p(\pi(q))$, then $p(\pi)>p(\pi\prime)$.
  \end{theorem}

  In other words, exchanging two positions in a permutation where the
  horse more likely to win is not ranked before the other would lead
  to an increase of the permutation probability. 

  This suggests the ground-truth permutation, ranked decreasingly by
  their probabilities, owns the maximum permutation probability on a
  given distribution. In SMT, we are motivated to optimize parameters
  to maximize the likelihood of ground-truth permutation of an N-best
  hypotheses.

  Due to the limitation of space, see \cite{plackett1975analysis,cao2007learning}
  for the proofs of the theorems.

  \section{Plackett-Luce Model in Statistical Machine Translation}

  In SMT, let $\mathbf{f}=(f_{1},f_{2}\ldots)$ denote source sentences,
  and $\mathbf{e}=(\{e_{1,1},\ldots\},\{e_{2,1},\ldots\}\ldots)$ denote
  target hypotheses. A set of features are defined on both source and
  target side. We refer to $h(e_{i,*})$ as a feature vector of a hypothesis
  from the $i$th source sentence, and its score from a ranking function
  is defined as the inner product $h(e_{i,*})^{T}\boldsymbol{w}$ of
  the weight vector $\boldsymbol{w}$ and the feature vector. 

  We first follow the popular exponential style to define a parameterized
  probability distribution over a list of hypotheses.

  \begin{equation}
    p(e_{i,j})=\frac{\exp\{h(e_{i,j})^{T}\boldsymbol{w}\}}{\sum_{k}\exp\{h(e_{i,k})^{T}\boldsymbol{w}\}}
  \end{equation}

  The ground-truth permutation of an $n$best list is simply obtained
  after ranking by their sentence-level BLEUs. Here we only concentrate
  on their relative ranks which are straightforward to compute in practice,
  e.g. add 1 smoothing. Let $\pi_{i}^{*}$ be the ground-truth permutation
  of hypotheses from the $i$th source sentences, and our optimization
  objective is maximizing the log-likelihood of the ground-truth permutations
  and penalized using a zero-mean and unit-variance Gaussian prior.
  This results in the following objective and gradient:

  \begin{equation}
    \mathcal{L}=\log\{\prod_{i}p(\pi_{i}^{*},\mathcal{P}(\mathbf{\boldsymbol{w}}))\}-\frac{1}{2}\boldsymbol{w}^{T}\boldsymbol{w}
  \end{equation}

  \begin{equation}
    \frac{\partial\mathcal{L}}{\partial\boldsymbol{w}}=\sum_{i}\sum_{j}\{h(e_{i,\pi_{i}^{*}(j)})-\sum_{t=j}(h(e_{i,\pi_{i}^{*}(t)})\cdot\frac{p(e_{i,\pi_{i}^{*}(t)})}{Z_{i,j}})\}-\boldsymbol{w}
  \end{equation}
  where $Z_{i,j}$ is defined as the $Z_{j}$ in Formula (1) of the
  $i$th source sentence.

  The log-likelihood function is smooth, differentiable, and concave
  with the weight vector $\boldsymbol{w}$, and its local maximal solution
  is also a global maximum. Iteratively selecting one parameter in $\alpha$
  for tuning in a line search style (or MERT style) could also converge
  into the global global maximum \cite{bertsekas1999nonlinear}. In
  practice, we use more fast limited-memory BFGS (L-BFGS) algorithm
  \cite{byrd1995limited}.

  \subsection*{N-best Hypotheses Resample}

  The log-likelihood of a Plackett-Luce model is not a strict upper
  bound of the BLEU score, however, it correlates with BLEU well in
  the case of rich features. The concept of ``rich'' is actually qualitative,
  and obscure to define in different applications. We empirically provide
  a formula to measure the richness in the scenario of machine translation.

  \begin{equation}
    r=\frac{\textrm{the size of features }}{\textrm{the average size of N-best lists}}\label{eq:empiral-r}
  \end{equation}
  The greater, the richer. In practice, we find a rough threshold of
  r is 5.

  In engineering, the size of an N-best list with unique hypotheses
  is usually less than several thousands. This suggests that, if features
  are up to thousands or more, the Plackett-Luce model is quite suitable
  here. Otherwise, we could reduce the size of N-best lists by sampling
  to make $r$ beyond the threshold.

  Their may be other efficient sampling methods, and here we adopt a
  simple one. If we want to $m$ samples from a list of hypotheses $\mathbf{e}$,
  first, the $\frac{m}{3}$ best hypotheses and the $\frac{m}{3}$ worst
  hypotheses are taken by their sentence-level BLEUs. Second, we sample
  the remaining hypotheses on distribution $p(e_{i})\propto\exp(h(e_{i})^{T}\boldsymbol{w})$,
  where $\mathbf{\boldsymbol{w}}$ is an initial weight from last iteration.

  \section{Evaluation}

  \begin{table}[h]
    \begin{centering}
      \begin{tabular}{l|c|c|c}
        & MT02(dev)  & MT04(\%)  & MT05(\%) \tabularnewline
        \hline 
        MERT  & 34.61 & 31.76 & 28.85\tabularnewline
        MIRA & 35.31 & 32.25 & 29.37\tabularnewline
        \hline 
        PL(1) & 34.20 & 31.70 & 28.90\tabularnewline
        PL(2) & 34.31 & 31.83 & 29.10\tabularnewline
        PL(3) & 34.39 & 32.05 & 29.20\tabularnewline
        PL(4) & 34.40 & 32.13 & 29.46+\tabularnewline
        PL(5) & 34.46 & 32.19+ & 29.42+\tabularnewline
        PL(6) & 34.37 & 32.16 & 29.30\tabularnewline
        PL(7) & 34.39 & 32.20+ & 29.32\tabularnewline
        PL(8) & 34.70 & 32.19+ & 29.10\tabularnewline
        PL(9) & 34.30 & 32.07 & 29.22\tabularnewline
        PL(10) & 34.30 & 32.14 & 29.19\tabularnewline
        \hline 
      \end{tabular}
      \par
    \end{centering}
    {\scriptsize{}\caption{PL($k$): Plackett-Luce model optimizing the ground-truth permutation
    with length $k$. The significant symbols (+ at 0.05 level) are compared
    with MERT. The bold font numbers signifies better results compared
    to M(1) system. }
    }{\scriptsize\par}

    {\scriptsize{}\label{tb:all-data}} 
  \end{table}
  We compare our method with MERT and MIRA\footnote{MIRA is from the open-source Moses \cite{koehn2007moses}}
  in two tasks, iterative training, and N-best list rerank. We do not
  list PRO \cite{hopkins2011tuning} as our baseline, as Cherry et al.\cite{cherry2012batch}
  have compared PRO with MIRA and MERT massively. 

  In the first task, we align the FBIS data (about 230K sentence pairs)
  with GIZA++, and train a 4-gram language model on the Xinhua portion
  of Gigaword corpus. A hierarchical phrase-based (HPB) model (Chiang,
  2007) is tuned on NIST MT 2002, and tested on MT 2004 and 2005. All
  features are eight basic ones \cite{chiang2007hierarchical} and extra
  220 group features. We design such feature templates to group grammars
  by the length of source side and target side, (feat-type,a\ensuremath{\le}src-side\ensuremath{\le}b,c\ensuremath{\le}tgt-side\ensuremath{\le}d),
  where the feat-type denotes any of the relative frequency, reversed
  relative frequency, lexical probability and reversed lexical probability,
  and {[}a, b{]}, {[}c, d{]} enumerate all possible subranges of {[}1,
  10{]}, as the maximum length on both sides of a hierarchical grammar
  is limited to 10. There are 4 $\times$ 55 extra group features. 

  In the second task, we rerank an N-best list from a HPB system with
  7491 features from a third party. The system uses six million parallel
  sentence pairs available to the DARPA BOLT Chinese-English task. This
  system includes 51 dense features (translation probabilities, provenance
  features, etc.) and up to 7440 sparse features (mostly lexical and
  fertility-based). The language model is a 6-gram model trained on
  a 10 billion words, including the English side of our parallel corpora
  plus other corpora such as Gigaword (LDC2011T07) and Google News.
  For the tuning and test sets, we use 1275 and 1239 sentences respectively
  from the LDC2010E30 corpus. 

  \begin{figure*}
    \begin{centering}
      \includegraphics[scale=0.32]{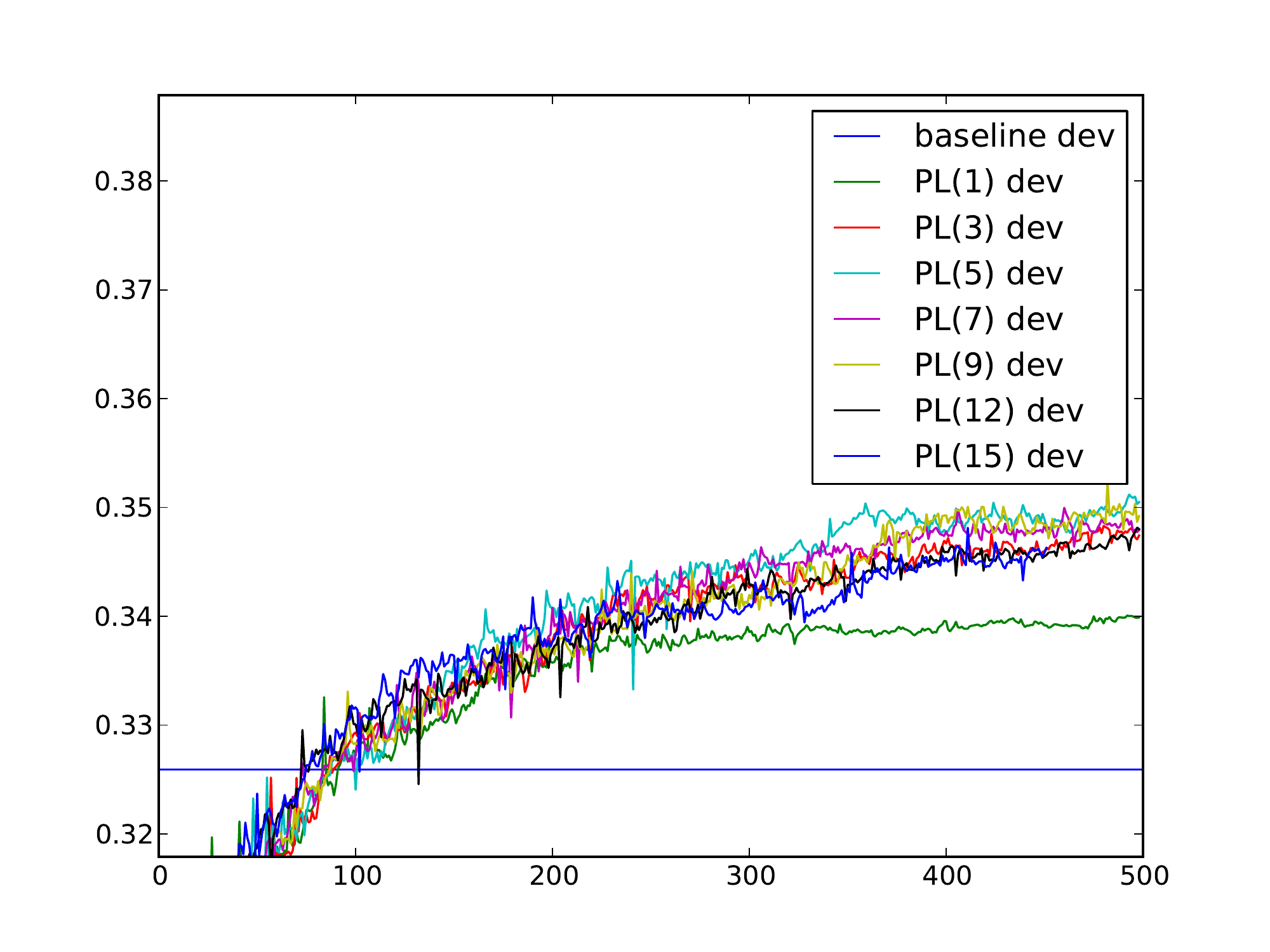}\includegraphics[scale=0.32]{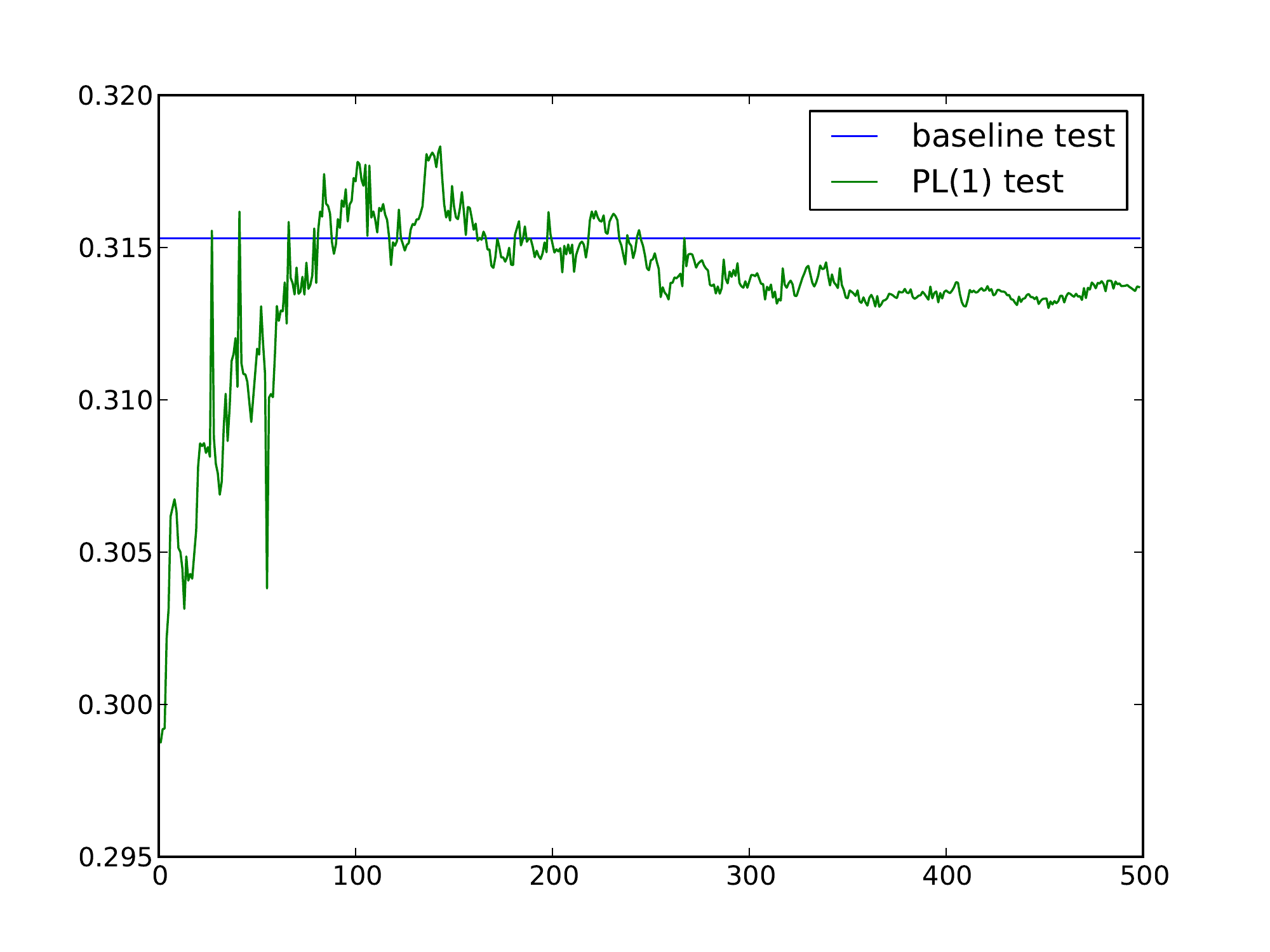}
      \par
    \end{centering}
    \begin{centering}
      \includegraphics[scale=0.32]{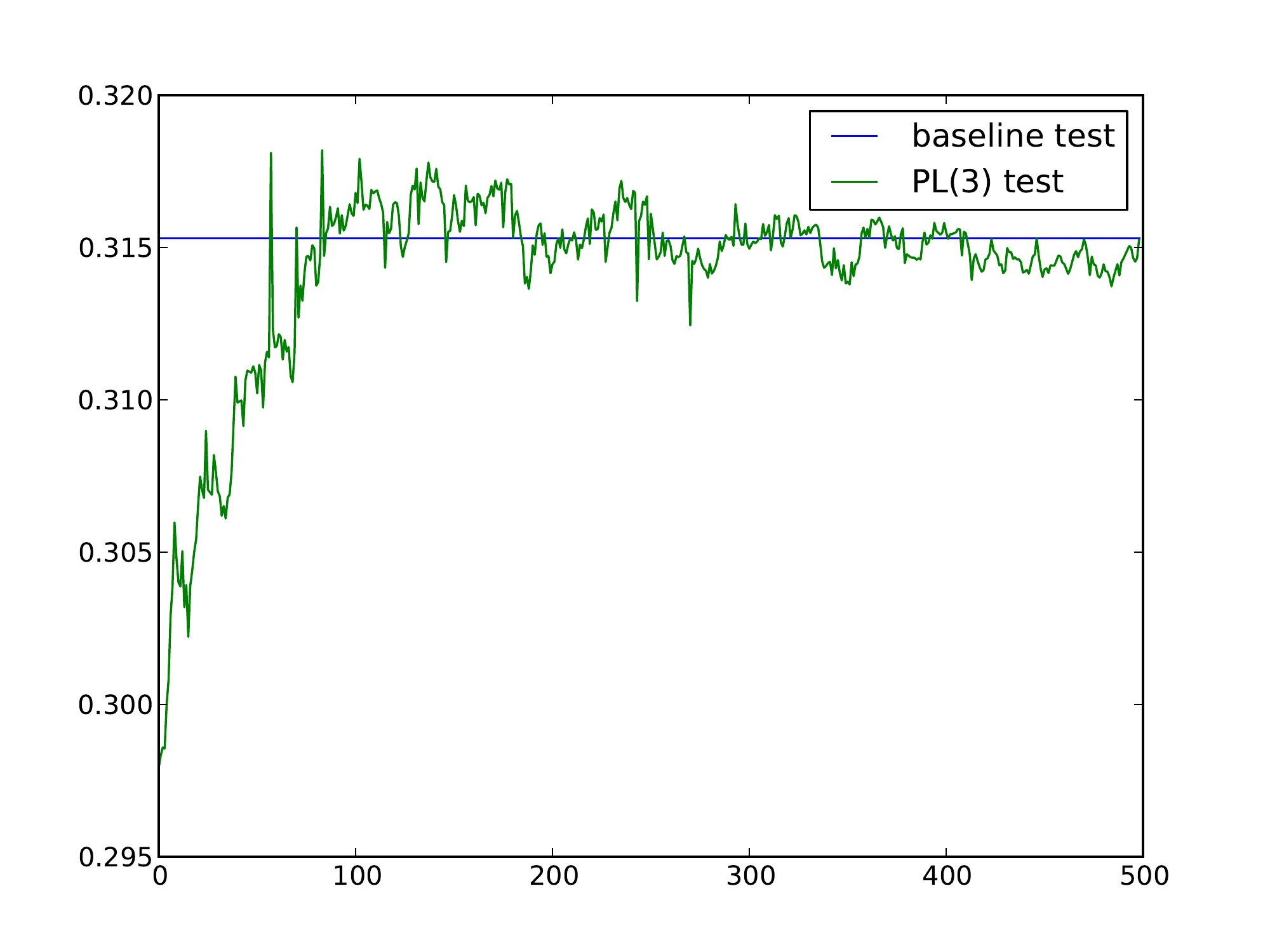}\includegraphics[scale=0.32]{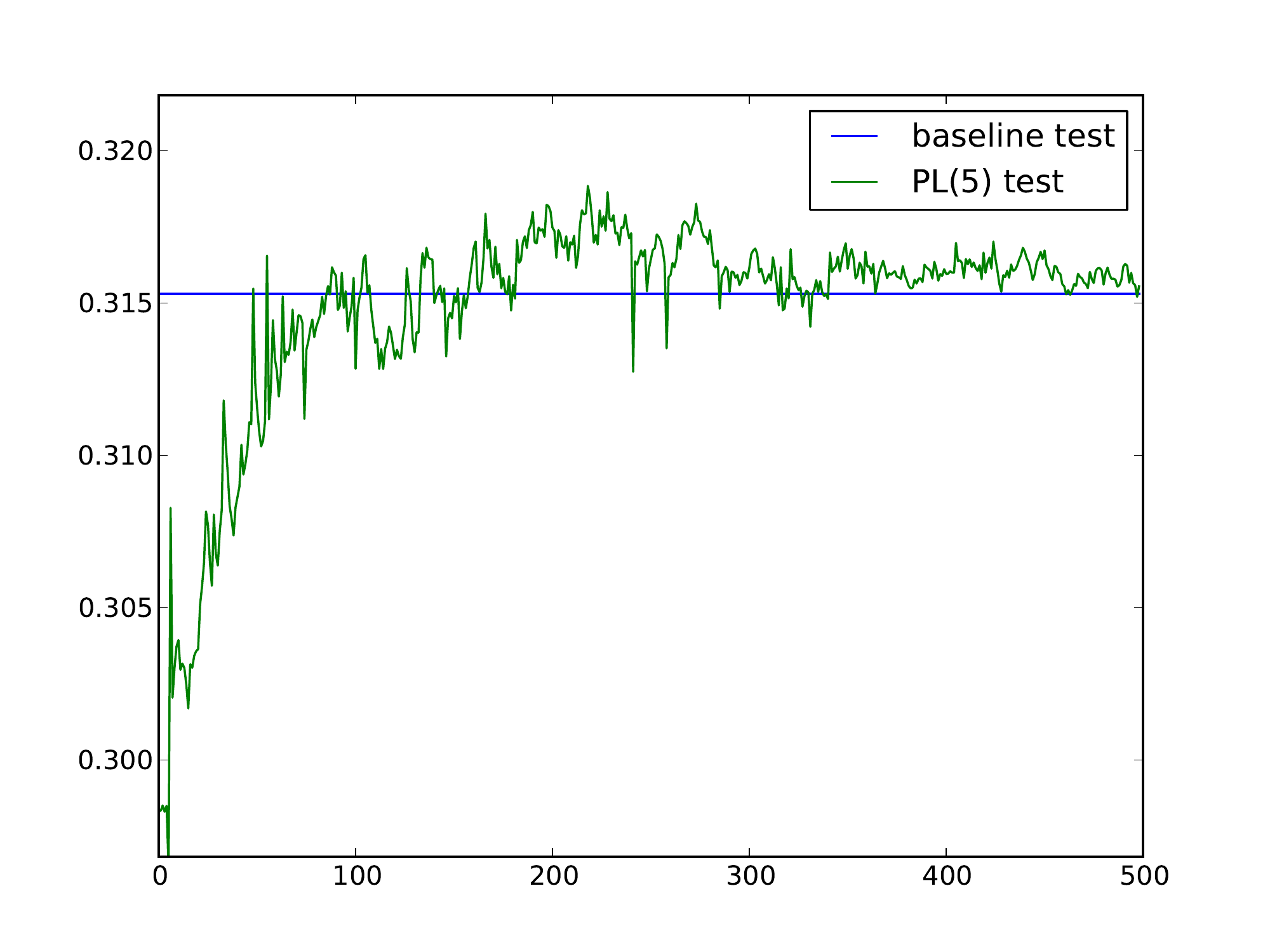}
      \par
    \end{centering}
    \begin{centering}
      \includegraphics[scale=0.32]{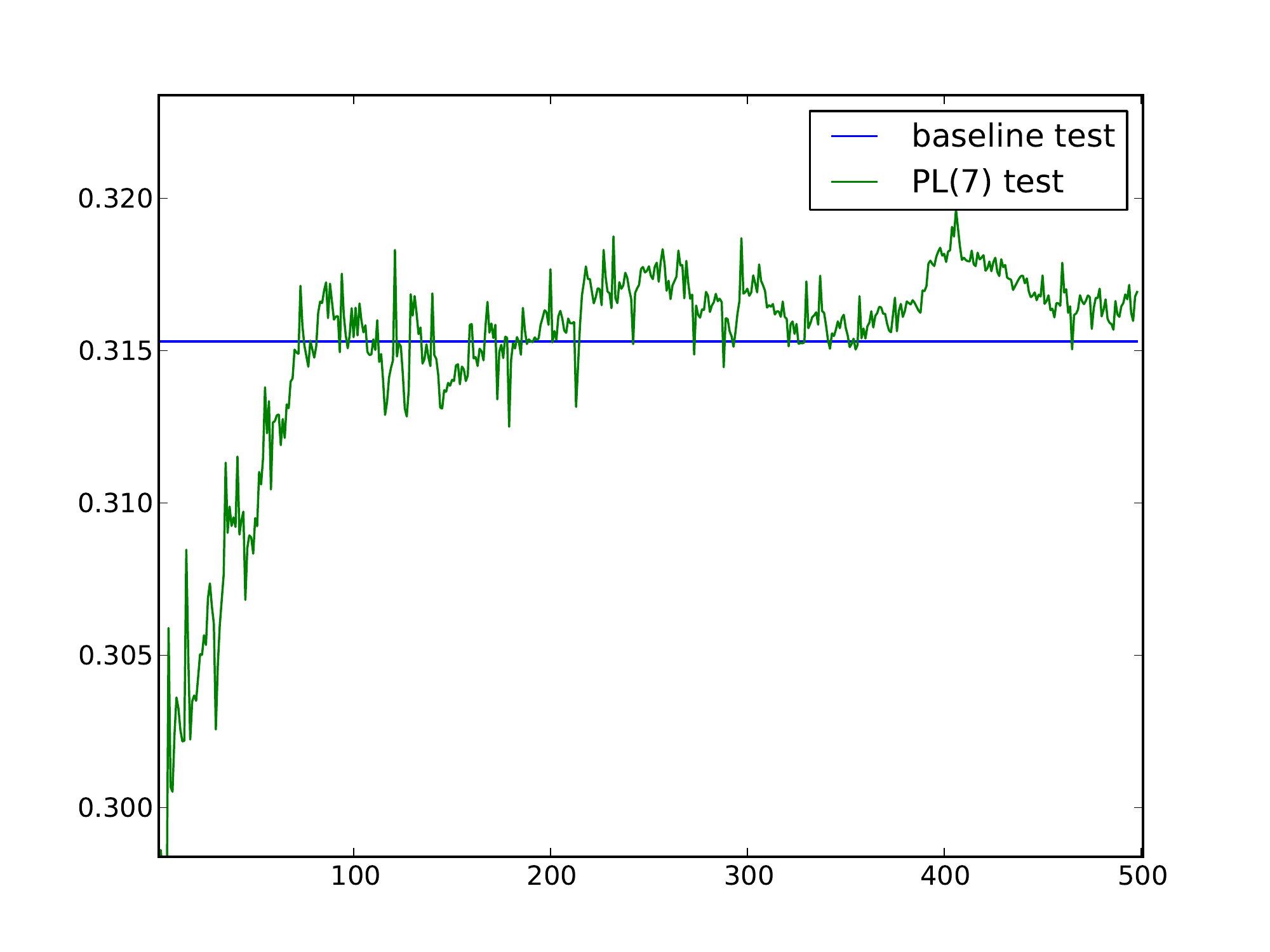}\includegraphics[scale=0.32]{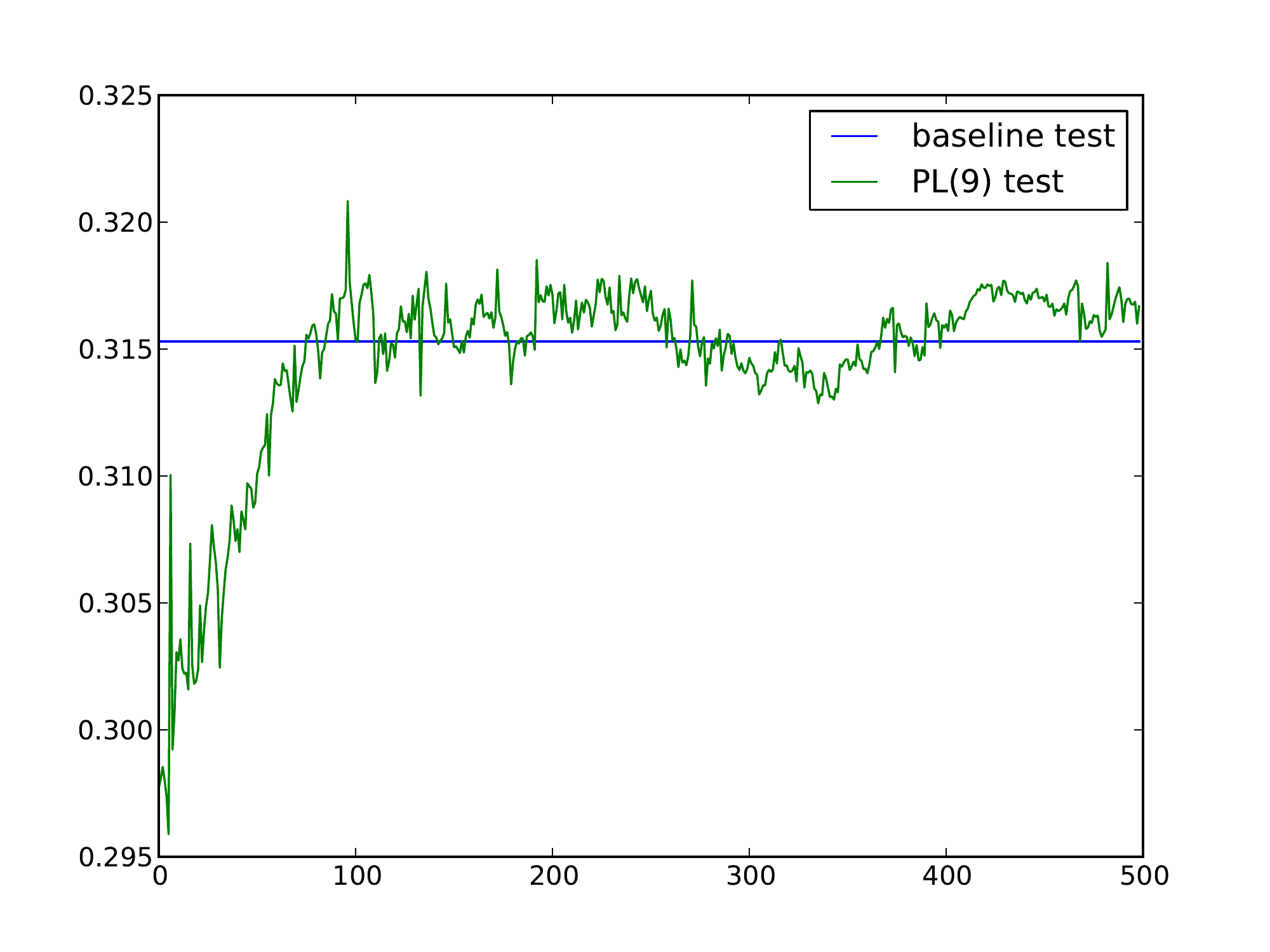}
      \par
    \end{centering}
    \begin{centering}
      \includegraphics[scale=0.32]{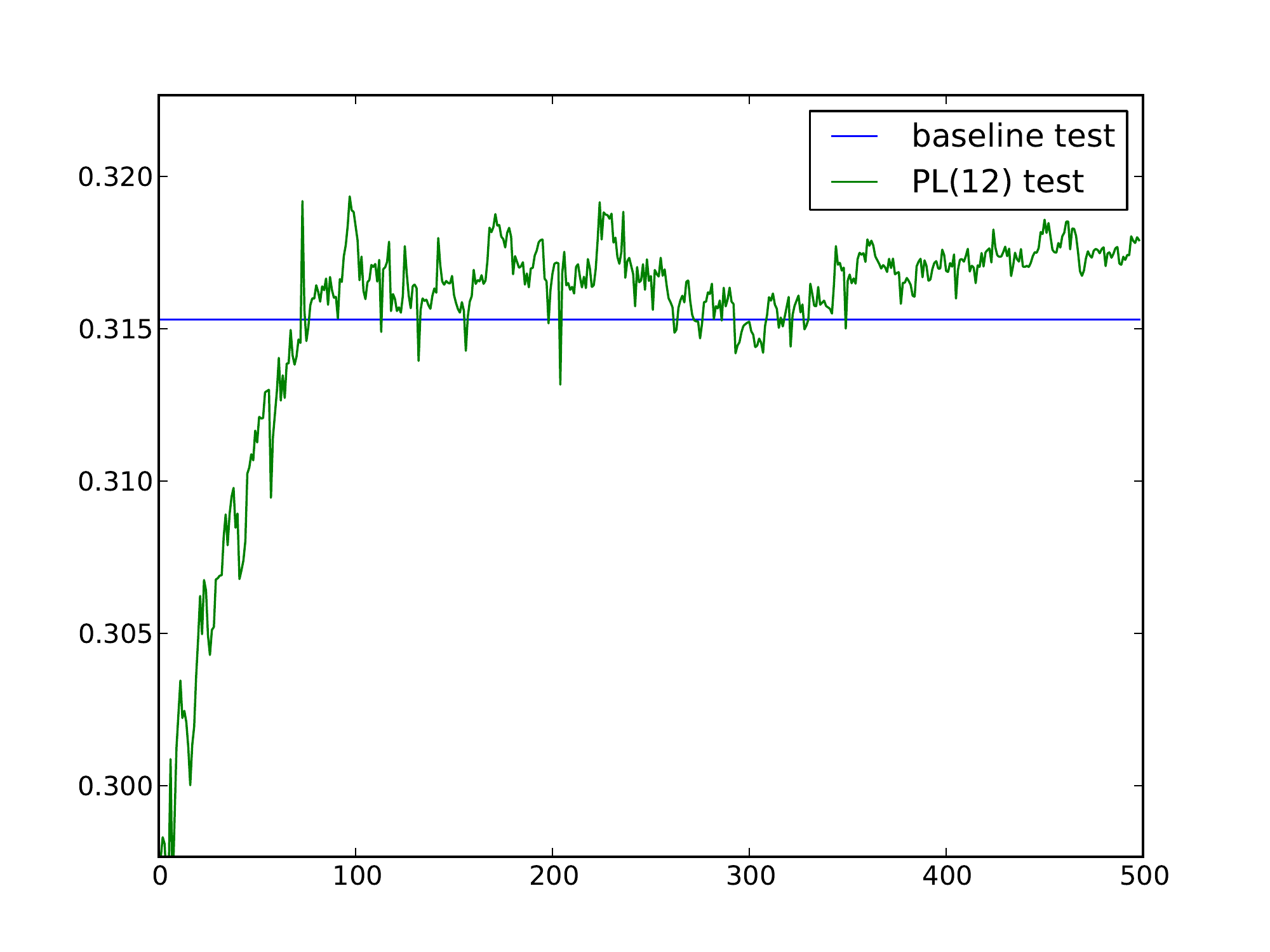}\includegraphics[scale=0.32]{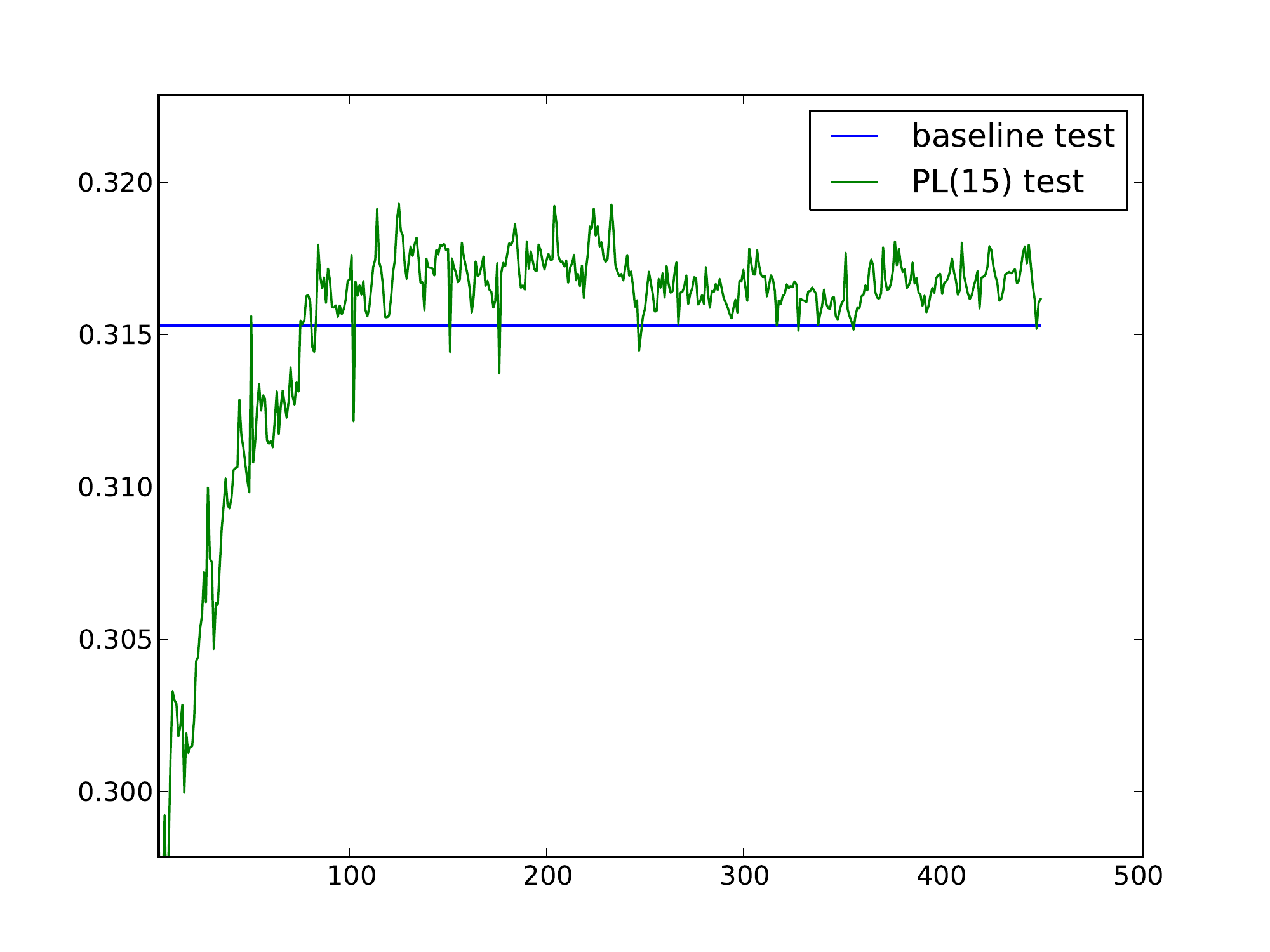}
      \par
    \end{centering}
    \caption{PL($k$) with 500 L-BFGS iterations, k=1,3,5,7,9,12,15 compared with
    MIRA in reranking.}
    \label{Fig:rerank}
  \end{figure*}

  \subsection{Plackett-Luce Model for SMT Tuning}

  We conduct a full training of machine translation models. By default,
  a decoder is invoked for at most 40 times, and each time it outputs
  $200$ hypotheses to be combined with those from previous iterations
  and sent into tuning algorithms. 

  In getting the ground-truth permutations, there are many ties with
  the same sentence-level BLEU, and we just take one randomly. In this
  section, all systems have only around two hundred features, hence
  in Plackett-Luce based training, we sample 30 hypotheses in an accumulative
  $n$best list in each round of training.

  All results are shown in Table \ref{tb:all-data}, we can see that
  all PL($k$) systems does not perform well as MERT or MIRA in the
  development data, this maybe due to that PL($k$) systems do not optimize
  BLEU and the features here are relatively not enough compared to the
  size of N-best lists (empirical Formula \ref{eq:empiral-r}). However,
  PL($k$) systems are better than MERT in testing. PL($k$) systems
  consider the quality of hypotheses from the $2$th to the $k$th,
  which is guessed to act the role of the margin like SVM in classification
  . Interestingly, MIRA wins first in training, and still performs quite
  well in testing.

  The PL(1) system is equivalent to a max-entropy based algorithm \cite{och2002discriminative}
  whose dual problem is actually maximizing the conditional probability
  of one oracle hypothesis. When we increase the $k$, the performances
  improve at first. After reaching a maximum around $k=5$, they decrease
  slowly. We explain this phenomenon as this, when features are rich
  enough, higher BLEU scores could be easily fitted, then longer ground-truth
  permutations include more useful information. 

  \subsection{Plackett-Luce Model for SMT Reranking}

  After being de-duplicated, the N-best list has an average size of
  around 300, and with 7491 features. Refer to Formula \ref{eq:empiral-r},
  this is ideal to use the Plackett-Luce model. Results are shown in
  Figure \ref{Fig:rerank}. We observe some interesting phenomena.

  First, the Plackett-Luce models boost the training BLEU very greatly,
  even up to 2.5 points higher than MIRA. This verifies our assumption,
  richer features benefit BLEU, though they are optimized towards a
  different objective.

  Second, the over-fitting problem of the Plackett-Luce models PL($k$)
  is alleviated with moderately large $k$. In PL(1), the over-fitting
  is quite obvious, the portion in which the curve overpasses MIRA is
  the smallest compared to other $k$, and its convergent performance
  is below the baseline. When $k$ is not smaller than 5, the curves
  are almost above the MIRA line. After 500 L-BFGS iterations, their
  performances are no less than the baseline, though only by a small
  margin.

  This experiment displays, in large-scale features, the Plackett-Luce
  model correlates with BLEU score very well, and alleviates overfitting
  in some degree.

  \bibliographystyle{plain}
  \nocite{*}
  \bibliography{references}

\begin{thebibliography}{10}

\bibitem{berger1996maximum}
Adam~L Berger, Vincent J~Della Pietra, and Stephen A~Della Pietra.
\newblock A maximum entropy approach to natural language processing.
\newblock {\em Computational linguistics}, 22(1), 1996.

\bibitem{bertsekas1999nonlinear}
Dimitri~P Bertsekas.
\newblock Nonlinear programming.
\newblock 1999.

\bibitem{blunsom2008discriminative}
Phil Blunsom, Trevor Cohn, and Miles Osborne.
\newblock A discriminative latent variable model for statistical machine
  translation.
\newblock In {\em Proc. of ACL}, 2008.

\bibitem{byrd1995limited}
Richard~H Byrd, Peihuang Lu, Jorge Nocedal, and Ciyou Zhu.
\newblock A limited memory algorithm for bound constrained optimization.
\newblock {\em SIAM Journal on Scientific Computing}, 16(5), 1995.

\bibitem{cao2007learning}
Zhe Cao, Tao Qin, Tie-Yan Liu, Ming-Feng Tsai, and Hang Li.
\newblock Learning to rank: from pairwise approach to listwise approach.
\newblock In {\em Proc. of ICML}, 2007.

\bibitem{cer2008regularization}
Daniel Cer, Daniel Jurafsky, and Christopher~D Manning.
\newblock Regularization and search for minimum error rate training.
\newblock In {\em Proc. of WSMT}, 2008.

\bibitem{cherry2012batch}
Colin Cherry and George Foster.
\newblock Batch tuning strategies for statistical machine translation.
\newblock In {\em Proceedings of the 2012 Conference of the North American
  Chapter of the Association for Computational Linguistics: Human Language
  Technologies}, pages 427--436. Association for Computational Linguistics,
  2012.

\bibitem{chiang2005hierarchical}
David Chiang.
\newblock A hierarchical phrase-based model for statistical machine
  translation.
\newblock In {\em Proc. of ACL}, 2005.

\bibitem{chiang2007hierarchical}
David Chiang.
\newblock Hierarchical phrase-based translation.
\newblock {\em computational linguistics}, 33(2), 2007.

\bibitem{chiang2008online}
David Chiang, Yuval Marton, and Philip Resnik.
\newblock Online large-margin training of syntactic and structural translation
  features.
\newblock In {\em Proc. of EMNLP}, 2008.

\bibitem{collins2002discriminative}
Michael Collins.
\newblock Discriminative training methods for hidden markov models: Theory and
  experiments with perceptron algorithms.
\newblock In {\em Proc. of EMNLP}, 2002.

\bibitem{duh2008beyond}
Kevin Duh and Katrin Kirchhoff.
\newblock Beyond log-linear models: boosted minimum error rate training for
  n-best re-ranking.
\newblock In {\em Proc. of ACL: Short Papers}, 2008.

\bibitem{galley2011optimal}
Michel Galley and Chris Quirk.
\newblock Optimal search for minimum error rate training.
\newblock In {\em Proc. of EMNLP}, 2011.

\bibitem{galley2013regularized}
Michel Galley, Chris Quirk, Colin Cherry, and Kristina Toutanova.
\newblock Regularized minimum error rate training.
\newblock In {\em EMNLP}, pages 1948--1959, 2013.

\bibitem{hopkins2011tuning}
Mark Hopkins and Jonathan May.
\newblock Tuning as ranking.
\newblock In {\em Proc. of EMNLP}, 2011.

\bibitem{koehn2007moses}
Philipp Koehn, Hieu Hoang, Alexandra Birch, Chris Callison-Burch, Marcello
  Federico, Nicola Bertoldi, Brooke Cowan, Wade Shen, Christine Moran, Richard
  Zens, et~al.
\newblock Moses: Open source toolkit for statistical machine translation.
\newblock In {\em Proc. of ACL: Poster}, 2007.

\bibitem{koehn2003statistical}
Philipp Koehn, Franz~Josef Och, and Daniel Marcu.
\newblock Statistical phrase-based translation.
\newblock In {\em Proc. of NAACL}, 2003.

\bibitem{kumar2004minimum}
Shankar Kumar and William~J Byrne.
\newblock Minimum bayes-risk decoding for statistical machine translation.
\newblock In {\em Proc. of HLT-NAACL}, 2004.

\bibitem{kumar2009efficient}
Shankar Kumar, Wolfgang Macherey, Chris Dyer, and Franz Och.
\newblock Efficient minimum error rate training and minimum bayes-risk decoding
  for translation hypergraphs and lattices.
\newblock In {\em Proc. of Joint ACL and AFNLP}, 2009.

\bibitem{lafferty2001conditional}
John Lafferty, Andrew McCallum, and Fernando~CN Pereira.
\newblock Conditional random fields: Probabilistic models for segmenting and
  labeling sequence data.
\newblock In {\em Prof. of ICML}, 2001.

\bibitem{li2009first}
Zhifei Li and Jason Eisner.
\newblock First-and second-order expectation semirings with applications to
  minimum-risk training on translation forests.
\newblock In {\em Proc. of EMNLP}, 2009.

\bibitem{liang2006end}
Percy Liang, Alexandre Bouchard-C{\^o}t{\'e}, Dan Klein, and Ben Taskar.
\newblock An end-to-end discriminative approach to machine translation.
\newblock In {\em Proc. of ACL}, 2006.

\bibitem{macherey2008lattice}
Wolfgang Macherey, Franz~Josef Och, Ignacio Thayer, and Jakob Uszkoreit.
\newblock Lattice-based minimum error rate training for statistical machine
  translation.
\newblock In {\em Proc. of EMNLP}, 2008.

\bibitem{mcdonald2005online}
Ryan McDonald, Koby Crammer, and Fernando Pereira.
\newblock Online large-margin training of dependency parsers.
\newblock In {\em Proc. of ACL}, 2005.

\bibitem{moore2008random}
Robert~C Moore and Chris Quirk.
\newblock Random restarts in minimum error rate training for statistical
  machine translation.
\newblock In {\em Proceedings of the 22nd International Conference on
  Computational Linguistics-Volume 1}, pages 585--592. Association for
  Computational Linguistics, 2008.

\bibitem{och2003minimum}
Franz~Josef Och.
\newblock Minimum error rate training in statistical machine translation.
\newblock In {\em Proc. of ACL}, 2003.

\bibitem{och2002discriminative}
Franz~Josef Och and Hermann Ney.
\newblock Discriminative training and maximum entropy models for statistical
  machine translation.
\newblock In {\em Proc. of ACL}, 2002.

\bibitem{pauls2009consensus}
Adam Pauls, John DeNero, and Dan Klein.
\newblock Consensus training for consensus decoding in machine translation.
\newblock In {\em Proc. of EMNLP}, 2009.

\bibitem{plackett1975analysis}
Robin~L Plackett.
\newblock The analysis of permutations.
\newblock {\em Applied Statistics}, 1975.

\bibitem{smith2006minimum}
David~A Smith and Jason Eisner.
\newblock Minimum risk annealing for training log-linear models.
\newblock In {\em Proc. of COLING/ACL: Poster}, 2006.

\bibitem{tan2013corpus}
Ming Tan, Tian Xia, Shaojun Wang, and Bowen Zhou.
\newblock A corpus level mira tuning strategy for machine translation.
\newblock In {\em EMNLP}, pages 851--856, 2013.

\bibitem{taskar2004max}
Ben Taskar, Dan Klein, Michael Collins, Daphne Koller, and Christopher~D
  Manning.
\newblock Max-margin parsing.
\newblock In {\em Proc. of EMNLP}, volume~1, 2004.

\bibitem{tillmann2006discriminative}
Christoph Tillmann and Tong Zhang.
\newblock A discriminative global training algorithm for statistical mt.
\newblock In {\em Proc. of ACL}, 2006.

\bibitem{tromble2008lattice}
Roy~W Tromble, Shankar Kumar, Franz Och, and Wolfgang Macherey.
\newblock Lattice minimum bayes-risk decoding for statistical machine
  translation.
\newblock In {\em Proc. of EMNLP}, 2008.

\bibitem{watanabe2007online}
Taro Watanabe, Jun Suzuki, Hajime Tsukada, and Hideki Isozaki.
\newblock Online large-margin training for statistical machine translation.
\newblock In {\em Proc. of EMNLP}, 2007.

\bibitem{yu2013max}
Heng Yu, Liang Huang, Haitao Mi, and Kai Zhao.
\newblock Max-violation perceptron and forced decoding for scalable mt
  training.
\newblock In {\em EMNLP}, pages 1112--1123, 2013.

\bibitem{zhu2001kernel}
Ji~Zhu and Trevor Hastie.
\newblock Kernel logistic regression and the import vector machine.
\newblock In {\em Proc. of NIPS}, 2001.

\end{thebibliography}

\end{document}